\documentclass[letterpaper, 10 pt, conference]{ieeeconf}
\IEEEoverridecommandlockouts                              
\usepackage[T1]{fontenc}

\usepackage[space, compress, sort]{cite}

\usepackage{amssymb} 
\usepackage{amsfonts}
\usepackage{amsthm}
\usepackage{wrapfig}
\usepackage[ruled,vlined]{algorithm2e}
\usepackage{mathtools}
\usepackage{tabularx}
\usepackage{graphicx}
\usepackage{enumerate}
\usepackage{float}
\usepackage{url}
\usepackage{verbatim}
\usepackage{booktabs} 
\usepackage{multirow}
\usepackage[linkcolor=black,citecolor=black,urlcolor=black,colorlinks=true]{hyperref}
\usepackage{bm}
\usepackage{subfigure}
\usepackage{algpseudocode}

\newtheorem{definition}{Definition}

\SetKwRepeat{Do}{do}{while}

\usepackage{xcolor}

\newcommand{\ie}{\textit{i.e.}}

\newcommand{\mc}[1]{{\mathcal{#1}}}

\title{\LARGE \bf  Resilient Multi-Robot Target Tracking with 
Sensing and Communication Danger Zones
}

\author{Peihan Li$^{1}$,Yuwei Wu$^{2}$, Jiazhen Liu$^{3}$, Gaurav S. Sukhatme$^{4}$, Vijay Kumar$^{2}$,  Lifeng Zhou$^{1}$\textsuperscript{\textdagger} 
\thanks{$^{1}$Peihan Li and Lifeng Zhou are with the Department of Electrical and Computer Engineering, Drexel University, Philadelphia, PA 19104, USA. Email: \texttt{\small \{pl525,lz457\}@drexel.edu}.}
\thanks{$^{2}$Yuwei Wu and Vijay Kumar are with the GRASP Lab, University of Pennsylvania, Philadelphia, PA 19104, USA. Email: \texttt{\small \{yuweiwu, kumar\}@seas.upenn.edu}.}
\thanks{$^{3}$Jiazhen Liu is with the Institute for Robotics and Intelligent Machines, Georgia Institute of Technology, Atlanta, GA 30332, USA. Email: \texttt{\small jliu3103@gatech.edu}.}
\thanks{$^{4}$Gaurav S. Sukhatme is with the Department of Computer Science, University of Southern California, Los Angeles, CA 90089, USA. Email:
\texttt{\small gaurav@usc.edu}.}
\thanks{This research was sponsored by the Army Research Lab through ARL DCIST CRA W911NF-17-2-0181.}
\thanks{\textsuperscript{\textdagger} Corresponding author.}
}

\begin{document}

\maketitle

\begin{abstract}

Multi-robot collaboration for target tracking in adversarial environments poses significant challenges, including system failures, dynamic priority shifts, and other unpredictable factors. These challenges become even more pronounced when the environment is unknown. In this paper, we propose a resilient coordination framework for multi-robot, multi-target tracking in environments with unknown sensing and communication danger zones. We consider scenarios where failures caused by these danger zones are probabilistic and temporary, allowing robots to escape from danger zones to minimize the risk of future failures. We formulate this problem as a nonlinear optimization with soft chance constraints, enabling real-time adjustments to robot behaviors based on varying types of dangers and failures. This approach dynamically balances target tracking performance and resilience, adapting to evolving sensing and communication conditions in real-time. To validate the effectiveness of the proposed method, we assess its performance across various tracking scenarios, benchmark it against methods without resilient adaptation and collaboration, and conduct several real-world experiments. 

\end{abstract}

\section{Introduction}

The multi-robot target tracking problem has been extensively studied for a wide range of applications, such as surveillance~\cite{rao1993fully}, environmental monitoring~\cite{dunbabin2012robots}, and wildfire coverage~\cite{8206579}.
Most of the research has focused on improving tracking accuracy, optimizing task allocation, and maximizing group efficiency within the constraints of limited resources~\cite{doi:10.1177/0278364917709507, 10160344, 9322567, li2023assignment, zhou2011multirobot, 4154838}.
However, deploying these methods in real-world environments often fails due to the inherent \textit{dangers} present in real-world scenarios.
The tracked targets might adversarially attack robots, and the environment could contain hazardous factors, both of which can cause the failure of task-critical resources and significantly degrade the tracking performance of the robot team. Mitigating these effects requires a thorough understanding of the associated risks, along with the development of resilient strategies against potential failures in planning the actions of robots.

To address the risk of failures during target tracking, \cite{liu2022decentralized, mayya2022adaptive, 9349130} developed several risk-aware tracking frameworks to balance maximizing team performance with minimizing the risk of sensor failures.
The authors in \cite{liu2024multi} extended the scenario by addressing the probabilistic communication failures.
In these studies, sensing danger zones interfere with the robots' sensors and cause sensor failures, while communication danger zones disrupt normal communication between robots, causing communication failures.
However, these studies focus on risk awareness in known zones and lack resilient coordination strategies that enable recovery and re-coordination when team members suffer from failures.

\begin{figure}[!t]
    \centering
    \subfigure[]{
\includegraphics[height = 3.8cm]{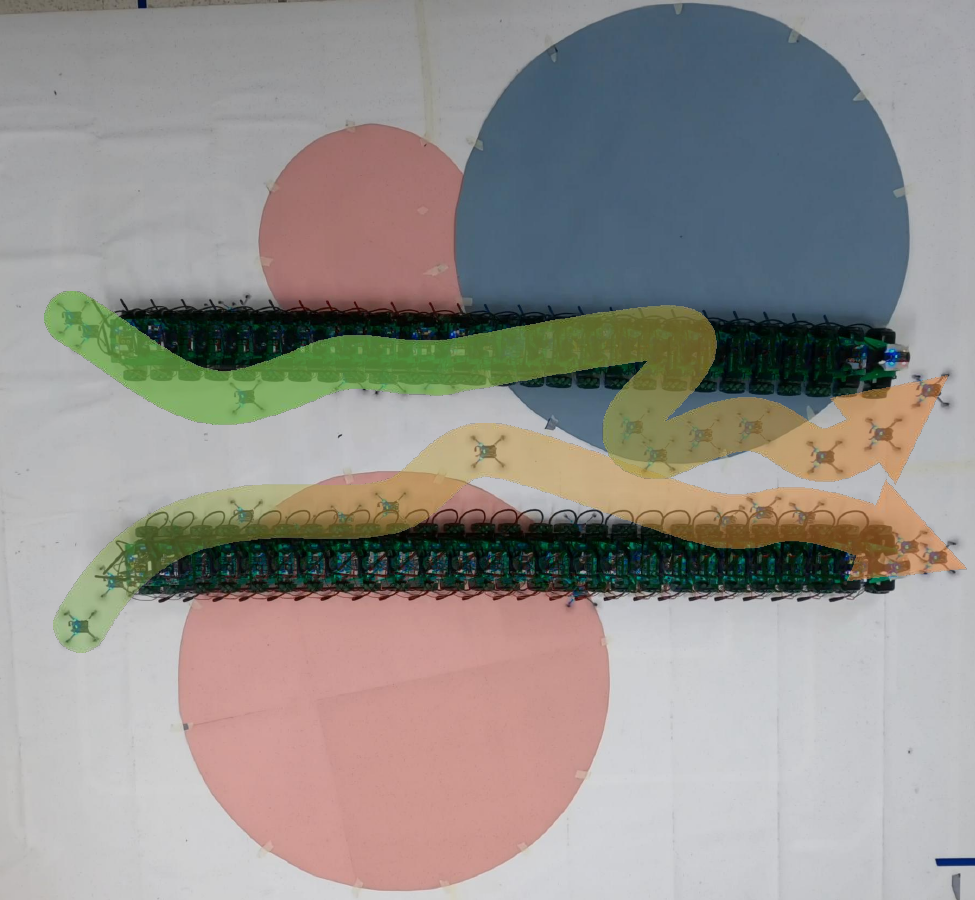}
    }
    \subfigure[]{
\includegraphics[height = 3.8cm]{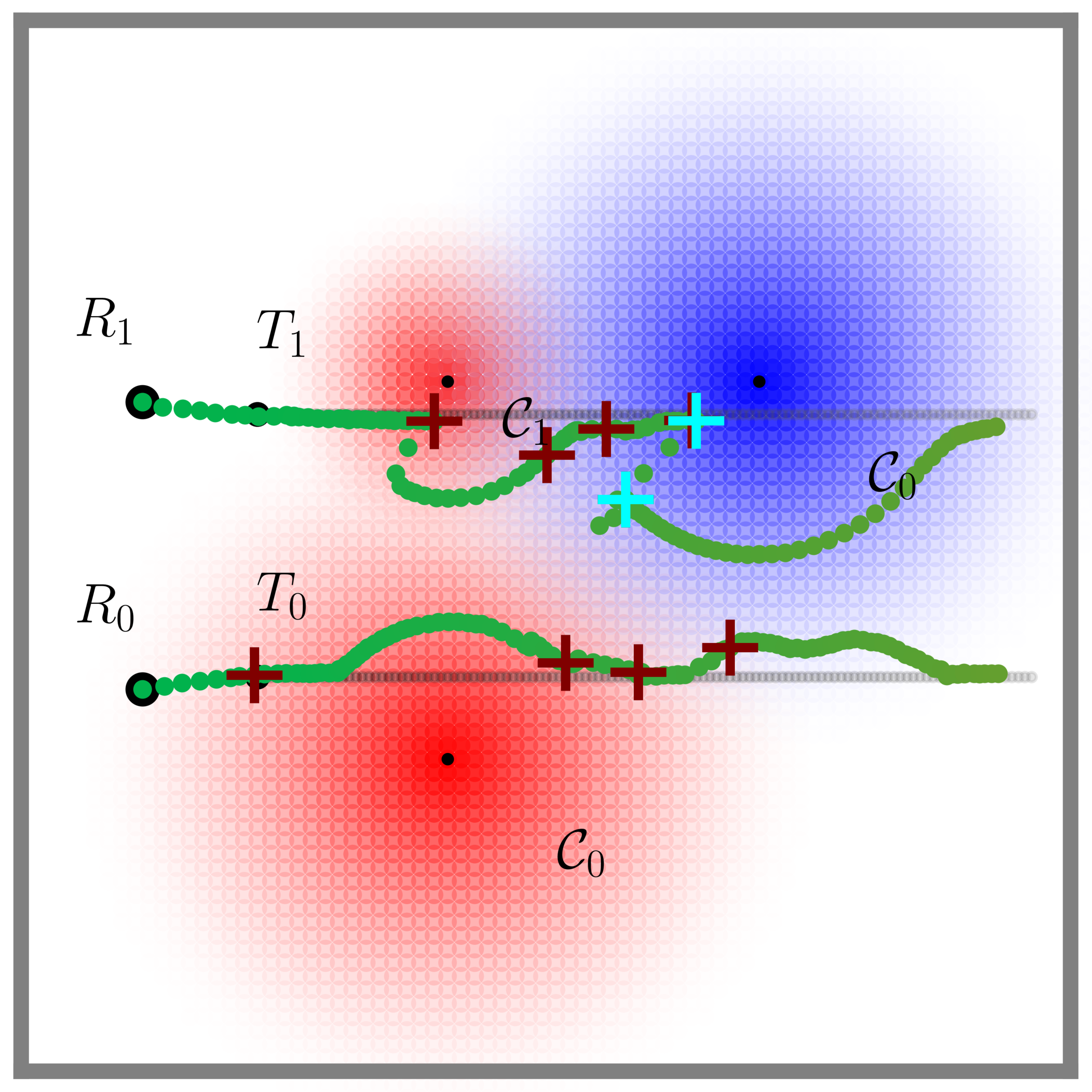}
    }
    \vspace{-0.2cm}
    \caption{Multi-robot target tracking with one communication danger zone (blue) and two sensing danger zones (red). (a) is the top view of the experiment, and (b) is a simulation with a similar setup. The red "+" represents a sensing attack, and the blue one represents a communication attack. The demo video is available at:~\url{https://youtu.be/0fq7fPuVaEA}.}
    \vspace{-0.5cm}
    \label{fig: fig1}
\end{figure}

To bridge these gaps, we propose a \textit{resilient} multi-robot target tracking framework for reactive task execution and recovery from failures that are caused by sensing and communication attacks. Our framework ensures resilience by enabling robots to recover from sensing and/or communication failures to sustain overall tracking performance. The main contributions of this paper are as follows:

\begin{itemize}
    \item We formulate multi-robot target tracking with \textit{initially unknown} danger zones as a partially centralized nonlinear optimization problem. The attack and recovery mechanisms are modeled for both sensing and communication danger zones.
    \item We propose a resilient coordination strategy. Attacked robots that have recovered can share information about the danger zones with their teammates, which enables the remaining robots to adjust their behaviors accordingly and reduce the likelihood of future failures.
    \item We extensively evaluate our framework through both simulations and hardware experiments (Fig.~\ref{fig: fig1}), demonstrating its resilience and adaptability.
\end{itemize}

\section{Related Works}

In real-world scenarios, tracking multiple targets requires robot teams to be collectively resilient to external attacks and adaptive to evolving environmental conditions. Prior works addressing these challenges can be grouped into two categories. The first focuses on improving robustness against attacks~\cite{mayya2022adaptive, liu2022decentralized, zhou2023robust, okumura2023fault,10008951, 9197243, liu2024multi, 9349130}, often via stealthy motion planning to proactively circumvent failures. For example,~\cite{mayya2022adaptive} balances tracking and safety against hostile targets;~\cite{liu2022decentralized} decentralizes this approach using consensus-based algorithms;~\cite{zhou2023robust} secures robots against any number of sensing and communication attacks.

The second category investigates resiliency, exploring how robots can quickly recover from failures, mitigate attacks, continue task execution, and maintain overall performance~\cite{prorok2021beyond, 8593630, 8534468, cavorsi2022multi, 10234549, 9354993, Abouelyazid_2023, schwager2017multi}. For example,~\cite{8593630} proposes an active information acquisition framework that can withstand any number of attacks and failures, even with minimal communication. Similarly,~\cite{8534468} develops a scalable resilient tracking strategy with approximation guarantees, while~\cite{schwager2017multi} presents an information gathering control policy under unknown hazards, enabling adaptation as robots gradually explore and learn hazard locations.

To further enhance resiliency, our prior work~\cite{Liresilient2024} proposed adaptation strategies that modify the problem formulation as new danger zones are discovered. Similarly,~\cite{mayya2022adaptive} achieves a balance between tracking performance and risk aversion by leveraging the covariance matrix for tracking accuracy and the observability Gramian for measuring risk. However, these approaches primarily consider offline planning with pre-defined environmental information, making them less effective for real-time adaptation in unknown or changing settings. Since dynamic environments require continual adjustment of planning strategies~\cite{8460863, 9561759}, our work introduces an optimization framework with soft chance constraints to enable flexible and resilient multi-robot behaviors in real time.

\section{Prerequisites}

\subsection{Models of Dynamics and Danger Zones}

We consider a team of robots, indexed as $\mathcal{R} = \{1, 2, \dots, M\}$, is tasked to track a team of targets indexed as $\mathcal{T} = \{1, 2, \dots, N\}$. 
Each robot can estimate the target's position relative to itself using a range and bearing sensor.
The environment contains $p$ sensing danger zones $ \{\mathcal{S}\}_{l=1}^{p} $ and $q$ communication danger zones $ \{\mathcal{C}\}_{k=1}^{q} $. 
The location of each danger zone remains unknown and is only revealed if a robot encounters an attack in it.
As soon as a robot enters a danger zone, it faces a certain probability of being attacked.
The attacked robot can share information about the position of the danger zone across the communication network with its teammates. We also assume the attack effect is temporary, so the attacked robot recovers after it exits the danger zone.
Inter-robot collision is not considered because robots operate at different altitudes.
Dynamics models of robots and targets, as well as details of danger zones, are introduced below. Notably, in this paper, we use \textit{attack} and \textit{failure} interchangeably. 

\begin{definition} (Robots and Targets Dynamics) Let $\mathbf{x}_{i, k} = [x_{i,t},~y_{i,t}]^\top$ denote the position of robot $i$, $i \in \mathcal{R}$ at time step $t$.
The discrete-time dynamics of the robot are defined as:
\vspace{-0.1cm}
\begin{equation}
    \mathbf{x}_{i, t+1} = \mathbf{\Phi}_i \mathbf{x}_{i, t} + \mathbf{\Lambda}_i \mathbf{u}_{i, t},
\end{equation}
where $\mathbf{\Phi}_i$ is the process matrix, $\mathbf{\Lambda}_i$ represents control matrix, and $\mathbf{u}_{i, t}$ is the control input. Similarly, the dynamics of target $j$, for any $j \in \mathcal{T}$, is
\vspace{-0.0cm}
\begin{equation}
    \mathbf{z}_{j, t+1} = \mathbf{A}_j \mathbf{z}_{j, t} + \mathbf{B}_j \mathbf{u}_{j, t},
\end{equation}
where $\mathbf{z}_{j, t} = [x_{T_{j,t}},~y_{T_{j, t}}]^\top$ denotes the target $j$'s position, $\mathbf{A}_j$ and $\mathbf{B}_j$ are the process and control matrix, respectively.
\end{definition}

\begin{definition}(Sensing Danger Zone) 
The sensing danger zone $\mathcal{S}_l$, built by the enemy, is defined by its center with position $\mathbf{x}_{\mathcal{S}_l}$, which follows a Gaussian distribution $\mathbf{x}_{\mathcal{S}_l}\sim\mathcal{N}(\mathbf{\mu}_{\mathcal{S}_l}, \mathbf{\Sigma}_{\mathcal{S}_l})$, and a safety clearance $r_l$. A robot at point $\mathbf{x}$ has a probability of $\phi$ to be attacked and lose tracking capability. The region within which a robot is at the risk of sensing attack is defined as:
\vspace{-0.0cm}
\begin{equation}
\mathcal{S}_l = \{\mathbf{x} \in \mathbb{R}^{n_x} :   \lVert \mathbf{x}_{\mathcal{S}_l} - \mathbf{x} \rVert \leq r_l \ \}.
 \end{equation}
The probability of the robot being within the sensing danger zone is:
\vspace{-0.0cm}
\begin{equation}
\begin{aligned}
    \mathbb{P}(\mathbf{x} \in \mathcal{S}_l) &= \int_{\lVert \mathbf{x}_{\mathcal{S}_l} - \mathbf{x} \rVert \leq r_l} \mathrm{pdf}(\mathbf{x}_{\mathcal{S}_l} - \mathbf{x})d(\mathbf{x}_{\mathcal{S}_l} - \mathbf{x}).\label{eq:integral_prob_typeI}
\end{aligned}
\end{equation}
\end{definition}
\begin{definition}(Communication Danger Zone) 
The communication danger zone $\mathcal{C}_k$ is assumed to be static and defined with its center position $\mathbf{x}_{\mathcal{C}_k}$ that follows a Gaussian distribution $\mathcal{N}(\mathbf{\mu}_{\mathcal{C}_k}, \mathbf{\Sigma}_{\mathcal{C}_k})$. 
The enemy in the danger zone can send noisy or deceiving signals to jam the robot, disabling its communication link. 
We use a similar definition as our prior work~\cite{liu2024multi} based on distance ratio to define the communication danger zone, inside which robot $i$'s communication with its teammate $j$ at $\mathbf{x}_j$ is at risk of being jammed when the following condition holds:
\vspace{-0.0cm}
\begin{equation}
\begin{aligned}
\mathcal{C}_k =\{ \mathbf{x}_i \in \mathbb{R}^{n_x} :     \lVert \mathbf{x}_{\mathcal{C}_k} - \mathbf{x}_i  \rVert  \leq  \delta_2  \lVert \mathbf{x}_i - \mathbf{x}_j  \rVert  \},
\label{eq: comm_define}
\end{aligned}
\end{equation}
where $\delta_2 \in \mathbb{R}_{>0}$ is a hyperparameter determined by the sensor characteristic. 
Hence, the probability of robot $i$ being jammed is 
\begin{equation}
\begin{aligned}
    \mathbb{P}(\mathbf{x}_i \in \mathcal{C}_k ) & = \int_{\lVert \mathbf{x}_{\mathcal{C}_k} - \mathbf{x}_i \rVert < \delta_2  \lVert \mathbf{x}_i - \mathbf{x}_j  \rVert } \hspace{-5.0mm} \mathrm{pdf}(\mathbf{x}_{\mathcal{C}_k} - \mathbf{x}_i )d(\mathbf{x}_{\mathcal{C}_k} - \mathbf{x}_i) .
    \label{eq:comm_prob_integrate} 
\end{aligned}
\end{equation}
Since a robot can communicate with multiple teammates, we calculate the highest probability of jamming in Eq.~\ref{eq:comm_prob_integrate} using distance from the farthest teammate, which we denote as $c^*_i$ for robot $i$.
\end{definition}
\vspace{-0.15cm}
We adopt Gaussian distributions to model the danger zones due to their mathematical tractability and their ability to naturally capture uncertainty about hazard locations. This continues, and probabilistic modeling enables analytical approximation of risk fields and facilitates real-time chance-constrained optimization.
\vspace{-0.15cm}
\subsection{Chance-constrained Optimization before Attacks}
In our previous study~\cite{liu2024multi}, we assumed prior knowledge of the danger zones and focused on robustness against potential attacks in target tracking. 
The tracking problem was formulated to ensure that all robots avoid danger zones while achieving the tracking tasks.
We can enforce an upper bound on the probability of entering any danger zone as chance constraints. 
We focus on the optimization problem to find the next step control for the robot team for optimal tracking, following the setup in~\cite{zhou2011multirobot}.
Then, the multi-robot tracking problem at timestamp $t$ is formulated as:
\vspace{-0.1cm}
\begin{subequations}
    \label{eq:opt}
    \begin{align}
        & \min_{\mathbf{u}_{i, t}, \forall i \in \mathcal{R}} \ \    w_1  \sum_{i=1}^{M} f(\mathbf{x}_{i, t+1}, \hat{\mathbf{z}}_{i,t+1}) + w_2 \sum_{i=1}^{M}\lVert \mathbf{u}_{i,t}\rVert \quad \ \ \ \label{eq:opt_objective}\\ 
        \textrm{s.t.} \
        & \mathbf{x}_{i, t+1} = \mathbf{\Phi}_i\mathbf{x}_{i,t} + \mathbf{\Lambda}_i\mathbf{u}_{i,t},  \forall i \in \mathcal{R}, \label{eq:opt_dynamics}\\
        &\mathbb{P}(\lVert \mathbf{x}_{\mathcal{S}_l} - \mathbf{x}_{i, t+1} \rVert \leq r_l) \leq \epsilon_1, \forall i \in \mathcal{R}, \ \forall l \in [p], \ \ \label{eq:opt_constraint_1}\\
         & \mathbb{P}(a_{ik} < \delta_2 c_{i}^*)\leq \epsilon_2, \forall i \in \mathcal{R}, \ \forall k \in [q], \  
 \label{eq:opt_constraint_2}
        \end{align}
\end{subequations}
where the objective function minimizes the weighted sum of tracking error in target state estimation $f(\cdot)$ and feasible control efforts with respective weights $w_1$ and $w_2$. The tracking error is a function of both robot states and the estimate of target positions, $\hat{\mathbf{z}}_{i, t+1}$, at time $t+1$. The $\epsilon_1, \epsilon_2 \in (0, 1)$ are the confidence levels for sensing and communication failures, respectively. 
Let the distance between robot $i$ to its farthest teammate be $c_{i}^*$, and the distance between robot $i$ and the jammer in the danger zone $\mathcal{C}_k$ be $a_{ik}$. 
The nonlinear optimization in Eq.~\ref{eq:opt} is approximated as:
\vspace{-0.1cm}
\begin{subequations}
    \label{eq:opt_approx}
    \begin{align}
        & \min_{\mathbf{u}_{i, t}, \forall i \in \mathcal{R} } \ \    w_1  \sum_{i=1}^{M} f(\mathbf{x}_{i, t+1}, \hat{\mathbf{z}}_{i,t+1}) + w_2\sum_{i=1}^{M}\lVert \mathbf{u}_{i,t}\rVert \quad  \label{eq:opt_approx_objective}\\ 
        \textrm{s.t.} \
        & \mathbf{x}_{i, t+1} = \mathbf{\Phi}_i\mathbf{x}_{i,t} + \mathbf{\Lambda}_i\mathbf{u}_{i,t},   \forall i \in \mathcal{R}, \label{eq:opt_approx_dynamics}\\
        &   \hat{\mathbf{a}}_{i, \mathcal{S}_l}^\top \mathbf{a}_{i, \mathcal{S}_l}  - r_l \geq \\
        & \quad \ \text{erf}^{-1}(1-2\epsilon_1)\sqrt{2\hat{\mathbf{a}}_{i, \mathcal{S}_l}^\top\mathbf{\Sigma}_{\mathcal{S}_l}\hat{\mathbf{a}}_{i, \mathcal{S}_l}} , \forall i \in \mathcal{R}, \forall l \in [p], \ \ \notag \label{eq:opt_approx_constraint_1}\\
         & \hat{\mathbf{a}}_{i, \mathcal{C}_k}^\top \mathbf{a}_{i, \mathcal{C}_k} - \delta_2 c^*  \geq \\
         & \quad  \text{erf}^{-1}(1 - 2\epsilon_2) \sqrt{2\hat{\mathbf{a}}_{i, \mathcal{C}_k}^\top \mathbf{\Sigma}_{\mathcal{C}_k}\hat{\mathbf{a}}_{i, \mathcal{C}_k}},  \forall i \in \mathcal{R},   \forall k \in [q], \notag
 \label{eq:opt_approx_constraint_2}
        \end{align}
\end{subequations}
where $\mathbf{a}_{i, \mathcal{S}_l} = \mathbf{\mu}_{\mathcal{S}_l}-\mathbf{x}_i$, $ \mathbf{a}_{i, \mathcal{C}_k} = \mathbf{\mu}_{\mathcal{C}_k} - \mathbf{x}_i$, and $ \hat{\mathbf{a}}_{i, \mathcal{S}_l},~ \hat{\mathbf{a}}_{i, \mathcal{C}_k}$ denote their corresponding normalized vectors.
$\text{erf}(\cdot)$ is the standard error function.

This problem formulation has a series of drawbacks: (i) It assumes that the prior locations of all danger zones are known beforehand; (ii) It adopts an over-conservative paradigm, in which robots never take risks for improved target tracking quality; (iii) No systematic recovery procedure is provided for robots to retain target tracking performance under adversarial attacks.

In this work, we address these limitations by assuming the environment is initially unknown, with danger zones gradually revealed as robots encounter attacks. These attacks allow the robots to explore the environment. Additionally, we design resilient strategies to help robots recover from failures after an attack.

\section{Resilient Target Tracking}

In our resilient target tracking framework, robots adopt different strategies to cope with sensing and communication attacks. 
Before encountering any attack from danger zones, the entire robot team initially has no prior knowledge of these zones and must be exposed to attacks to gather information on these zones.
The robots use onboard range and bearing sensors to estimate the targets' locations collectively.
The communication network transfers information instantaneously.
With instantaneous information shared within the communication network, robots' measurements are shared with all valid members, allowing the system to follow the optimal control strategy for maximizing tracking quality.
We assume that the communication network operates without range constraints.

Given that the entire robot team is indexed as $\mathcal{R}$, we use $\mathcal{R}^s$ to denote indices for the subset of robots with normal sensing capabilities and $\mathcal{R}^c$ to represent indices for the subset of robots connected in the communication network. In other words, robots whose indices belong to $\mathcal{R}^c$ form a \textit{communication group}.
Initially, both types of danger zones are unknown to all robots.
We let \( [p]^{d} \) denote indices for the set of sensing danger zones known to the robot performing individual planning, while \( [p]^{d, c} \) represents indices for the set of sensing danger zones shared in the communication group.
A similar definition applies to communication danger zones, where \( [q]^{d}\) represents indices for the set of communication danger zones known to the robot performing individual planning, and \( [q]^{d, c} \) corresponds to the set of communication danger zones shared within the communication group.

For robots in the communication group, a centralized framework is proposed to share information and collectively optimize tracking performance. 
Robots that receive only sensing attacks lose their sensing abilities but can still stay in the communication group for information sharing.
In contrast, when robots are under communication attacks or a combination of communication and sensing attacks, they need to perform individual planning and deploy different strategies.

\subsection{Danger Zone Attacks and Recovery Model}

We assume that the damage caused by danger zones is temporary and can be recovered under certain conditions. 
Robots can be attacked by both unknown and known danger zones.
To effectively simulate the attack and recovery processes, we model a risk field associated with each danger zone to represent probabilistic attacks. 
Additionally, we define corresponding recovery conditions when robots are under sensing or communication attacks.

\textbf{Sensing attacks}: we model the attack probability as:
\begin{equation}
\phi(\mathcal{S}_l) = \frac{1}{2 \pi | \mathbf{\Sigma}_{\mathcal{S}_l} | }\exp \left (-\frac{1}{2}\mathbf{a}_{i, \mathcal{S}_l}^\top  \mathbf{\Sigma}_{\mathcal{S}_l} \mathbf{a}_{i, \mathcal{S}_l}  \right ).
\end{equation}                                
Each sensing danger zone has an attack frequency at which it attempts to attack robots. 

\textbf{Sensing recovery}: 
In our problem setting, the attack is assumed to be temporary. 
If the probability of a robot being attacked is below a threshold $\mathbb{P}(\mathbf{x} \in \mathcal{S}_l) \leq \epsilon_{l,0}$, we assume the robot will recover from sensing failure. 
The threshold $\epsilon_{l,0}$ may differ from $\epsilon_1$ (see Eq.~\ref{eq:opt_constraint_1}), allowing for varied recovery conditions and simulating model mismatches. 
For instance, a more aggressive recovery behavior can be achieved if we make $\epsilon_{l, 0}$ smaller than $\epsilon_1$.

\textbf{Communication attacks}: The communication attack includes both the attack on the communication links between robots and a direct attack to jam all channels of a particular robot~\cite{10068300}.
We similarly model the communication danger zone $\mathcal{C}_k,~\forall k\in[q]$ as a risk field whose attack probability is
\begin{equation}
\phi(\mathcal{C}_k) = \frac{1}{2 \pi | \mathbf{\Sigma}_{\mathcal{C}_k} | }\exp \left (-\frac{1}{2} \mathbf{a}_{i, \mathcal{C}_k}^\top  \mathbf{\Sigma}_{\mathcal{C}_k} \mathbf{a}_{i, \mathcal{C}_k}  \right ).
\end{equation}
The communication danger zone continuously attacks the robots with a fixed frequency following the probability defined above. 
In addition, if the condition in Eq.~\ref{eq: comm_define} is satisfied, the robot will also directly get a communication attack that destroys its communication links.

\textbf{Communication recovery}: When the probability of jamming is no larger than an upper bound $\mathbb{P}(\mathbf{x}_i \in \mathcal{C}_k ) \leq \epsilon_{k,0}$, we assume it has recovered from the communication attacks.
The $\epsilon_{k,0}$ defines a probabilistic threshold for communication danger zones.
The attack frequency may vary in different danger zones to model their respective threat level.

\subsection{Target Tracking without Communication Attack}
If the current communication group is not empty, denoted as $\mathcal{R}^c \neq \varnothing $, we will conduct joint planning for robots within the communication group. 
Note that robots with communication ability may still face sensing attacks. In this case, the optimization problem is:
\vspace{-0.1cm}
\begin{subequations}
    \label{eq:opt_without_communication}
    \begin{align}
        & \min_{\mathbf{u}_{i,t} \nu_i, \xi_i \forall i \in \mathcal{R}^c } \ w_1 \sum_{i \in \mc{R}^s \cap \mathcal{R}^c} f(\mathbf{x}_{t+1}, \hat{\mathbf{z}}_{t+1})  + \\
        &\sum_{i \in\mathcal{R}^c} ( w_2 \lVert \mathbf{u}_{i,t}\rVert  + w_3 \sum_{\forall l \in [p]^{d, c}}  \| \nu_{i, l} \|  +  w_4 \sum_{\forall k \in [q]^{d}}  \| \xi_{i, k }\|  ) \quad \ \ \notag \\ 
        \textrm{s.t.} \
        & \mathbf{x}_{i, t+1} = \mathbf{\Phi}_i\mathbf{x}_{i, t} + \mathbf{\Lambda}_i\mathbf{u}_{i, t},  \forall  i \in\mathcal{R}^c,  \quad \\
        &    \hat{\mathbf{a}}_{i, \mathcal{S}_l}^\top \mathbf{a}_{i, \mathcal{S}_l} - r_l  + \nu_{i, l}    \geq \\
        & \quad \text{erf}^{-1}(1-2\epsilon_1)\sqrt{2\hat{\mathbf{a}}_{i, \mathcal{S}_l}^\top\mathbf{\Sigma}_{\mathcal{S}_l}\hat{\mathbf{a}}_{i, \mathcal{S}_l}} , \forall l \in [p]^{d, c}, i \in\mathcal{R}^c, \ \ \notag\\
         & \hat{\mathbf{a}}_{i, \mathcal{C}_k}^\top \mathbf{a}_{i, \mathcal{C}_k} - \delta_2 c^* +  \xi_{i, k} \geq \\
         & \quad \text{erf}^{-1}(1 - 2\epsilon_2) \sqrt{2\hat{\mathbf{a}}_{i, \mathcal{C}_k}^\top \mathbf{\Sigma}_{\mathcal{C}_k}\hat{\mathbf{a}}_{i, \mathcal{C}_k}},   \ \forall k \in [q]^{d, c}, i \in\mathcal{R}^c, \notag \\
        & \nu_{i, l}, \xi_{i, k} \in \mathbb{R}_{+}.
    \end{align}
\end{subequations}
The tracking error in the objective function is only for the robots with both sensing and communication abilities. $\nu_{i, l}$ and $\xi_{i, k}$ are slack variables that relax the constraints, allowing robots to resiliently take risks and enter danger zones for better tracking qualities. Weights $w_3$ and $w_4$ are selected empirically to balance tracking performance and risk avoidance. Larger values result in more conservative behaviors, while smaller values allow riskier actions for improved tracking.

\subsection{Resiliency and Recovery}
\label{sec:recovery_and_adaptiveness}

\subsubsection{Sensing Attacks}

Robots under sensing attacks are assumed to lose their sensing functionality but are still able to communicate with other robots.
When a sensing attack occurs, the attacked robots determine the probabilistic distribution of the danger zone's position, which is then broadcast to the rest of the team within the communication network instantly.
Let $\mathcal{S}^d$ be the sensing danger zones that have just been discovered.
Then, the set of known sensing danger zones revealed to the communication group is expanded, \ie~$[p]^{d, c} = [p]^{d, c}~\bigcup~\mathcal{S}^d$, and constraints in Eq.~\ref{eq:opt_without_communication} are adjusted accordingly. 
 
\subsubsection{Communication Attacks}

The robot under communication attack loses its communication capability and performs \textit{single-robot target tracking} and individual planning. 
It can track targets with its onboard measurements and update the estimate of the targets' positions.
The robot can also trace back to estimate the origin of the attack. 
However, such an estimate cannot be broadcast to the rest of the team immediately since the attacked robot's communication channel is disabled.  
Therefore, the task for a robot $r_i$ under communication attack would be to plan to maximize the tracking quality with its measurements and, at the same time, try to escape from the danger zone. After escaping from the danger zone and regaining communication, robot $r_i$ can broadcast the estimated location of the zone to the rest of the team within the communication network.

Robots under communication attacks estimate the positions of targets, optimize their trajectories to track targets, and independently navigate themselves out of the danger zone(s). 
Before the communication resumes, their teammates perform target tracking without knowing this communication danger zone.
The optimization problem for each robot $r_i$, $i \in \mathcal{R} \setminus \mathcal{R}^c$ under communication attack is formulated as: 
\vspace{-0.1cm} 
\begin{subequations}
\label{eq: single plan}
    \begin{align}
        & \min_{\mathbf{u}_{i,t},\nu,\xi} \ J (\mathbf{u}_{i,t},\nu,\xi )\\
        \textrm{s.t.} \
        & \mathbf{x}_{i, t+1} = \mathbf{\Phi}_i\mathbf{x}_{i, t} + \mathbf{\Lambda}_i\mathbf{u}_{i, t},  \quad \\
        &    \hat{\mathbf{a}}_{i, \mathcal{S}_l}^\top \mathbf{a}_{i, \mathcal{S}_l} - r_l  + \nu_l    \geq  \\
        & \qquad \text{erf}^{-1}(1-2\epsilon_1)\sqrt{2\hat{\mathbf{a}}_{i, \mathcal{S}_l}^\top\mathbf{\Sigma}_{\mathcal{S}_l}\hat{\mathbf{a}}_{i, \mathcal{S}_l}} , \forall l \in [p]^{d, c}, \ \ \notag \\
         & \hat{\mathbf{a}}_{i, \mathcal{C}_k}^\top \mathbf{a}_{i, \mathcal{C}_k} +  \xi_k \geq \label{eq: comm_single} \\
         & \qquad \text{erf}^{-1}(1 - 2\epsilon_2) \sqrt{2\hat{\mathbf{a}}_{i, \mathcal{C}_k}^\top \mathbf{\Sigma}_{\mathcal{C}_k}\hat{\mathbf{a}}_{i, \mathcal{C}_k}},   \ \forall k \in [q]^{d}, \notag \\
        &  \nu_{l}, \xi_{k} \in \mathbb{R}_{+}. 
    \end{align}
\end{subequations}
In the case of single-robot planning, the constraint of the communication danger zone  (\ref{eq: comm_single}) has $c^* = 0$.
Then, the objective function is defined as: 
\vspace{-0.1cm} 
\begin{align}
 J(\mathbf{u}_{i,t}, \nu, \xi ) & =     w_1  f(\mathbf{x}_{t+1}, \hat{\mathbf{z}}_{t+1}) + w_2 \lVert \mathbf{u}_{i,t}\rVert  + \\
        & w_3 \sum_{\forall l \in [p]^{d, c}}  \| \nu_l \| +  w_4 \sum_{\forall k \in [q]^{d}}  \| \xi_k \|.  \quad \ \  \notag
\end{align}
When the robot has recovered from the communication danger zone, information about the zone is broadcast to the rest of the team. We augment the set of known communication danger zones as $[q]^{d, c} = [q]^{d, c}\ \bigcup\ [q]^d$, supposing $[q]^d$ denotes indices for the set of communication danger zones which have been newly discovered.

\subsubsection{Sensing and Communication Attacks}

Consider a robot that is located in an overlapping region of sensing and communication danger zones and is thus subject to attacks from multiple sources.
Without any measurement or communication ability, the mission of such a robot would be to perform single-robot planning to escape from the dangerous region as soon as possible.
Its single-robot planning is done by solving an optimization similar to Eq.~\ref{eq: single plan}, except that its objective function is revised as:
\vspace{-0.1cm} 
\begin{align}
J(\mathbf{u}_{i,t}, \nu, \xi ) = &w_2\lVert \mathbf{u}_{i,t}\rVert  + \notag \\
&  w_3 \sum_{\forall l \in [p]^{d, c}}  \| \nu_l \|  +  w_4 \sum_{\forall k \in [q]^{d}}  \| \xi_k \|  ,
\end{align}
\ie, to escape from the dangerous region with minimum control effort. 
In addition, the weights for different danger zones can be adjusted so that the robot can prioritize the zone(s) which it is already in.

\subsection{Estimation Update and Recover}

We employ the \textit{trace} of the estimation covariance matrix to evaluate tracking performance.
Each robot is equipped with a range and a bearing sensor to measure all targets. 
Following the approach in~\cite{liu2024multi}, a standard Extended Kalman Filter (EKF) can be employed to update the estimates collectively for robots within the communication network.

As we have described in Sec.~\ref{sec:recovery_and_adaptiveness}, if a robot is under communication attack, it will perform single-robot target tracking using its onboard sensing while attempting to escape from the danger region since it is not part of the communication network and thus cannot exchange measurements with its teammates. 
In this case, the robot under attack and the rest of the team within the communication network run two separate EKFs to estimate the state of the targets. 
As soon as the robot successfully escapes from the communication danger zone and regains communication, it broadcasts its estimate to teammates in the communication network. 
To merge estimates from the two EKFs, we use \textit{covariance intersection} (CI). 
Let the estimation as well as covariance matrix from the robot under communication attack be $\hat{\mathbf{z}}_1$ and $\mathbf{\Sigma}_1$, the estimation and covariance matrix from the robots within the communication networks be $\hat{\mathbf{z}}_2$ and $\mathbf{\Sigma}_2$. 
Then, CI performs the following operations to get the \textit{merged} estimate $\hat{\mathbf{z}}_{m}$ and the \textit{merged} covariance matrix $\mathbf{\Sigma}_m$:
\vspace{-0.1cm} 
\begin{equation}
    \mathbf{\Sigma}^{-1}_{m} = \omega \mathbf{\Sigma}^{-1}_{1} + (1 - \omega) \mathbf{\Sigma}^{-1}_{2} 
\end{equation}
\begin{equation}
    \hat{\mathbf{z}}_{m} = \mathbf{\Sigma}_m (\omega \mathbf{\Sigma}^{-1}_{1}\hat{\mathbf{z}}_1 + (1-\omega)\mathbf{\Sigma}^{-1}_{2}\hat{\mathbf{z}}_{2})
\end{equation}
where $\omega$ is a parameter, and its value can be obtained by optimizing a particular norm~\cite{julier2007using}. After CI, all robots have the same estimate of targets, i.e., ($\hat{\mathbf{z}}_{m}$, $\mathbf{\Sigma}_m$).

\section{Results}

\subsection{Performance Analysis}

We conduct a series of simulation experiments involving sensing attacks, communication attacks, and their combination to evaluate the performance of our resilient target tracking framework. Each robot can observe and track multiple targets. Our approach is compared to a vanilla target tracking framework that lacks resilient recovery mechanisms against attacks.
More scenarios are available in our open-source code: \url{https://github.com/Zhourobotics/resilient-target-tracking.git}.

\subsubsection{Tracking under Sensing Attack}
\begin{figure}[!ht]
    \vspace{-0.1cm} 
    \centering
    \includegraphics[width=1\columnwidth]{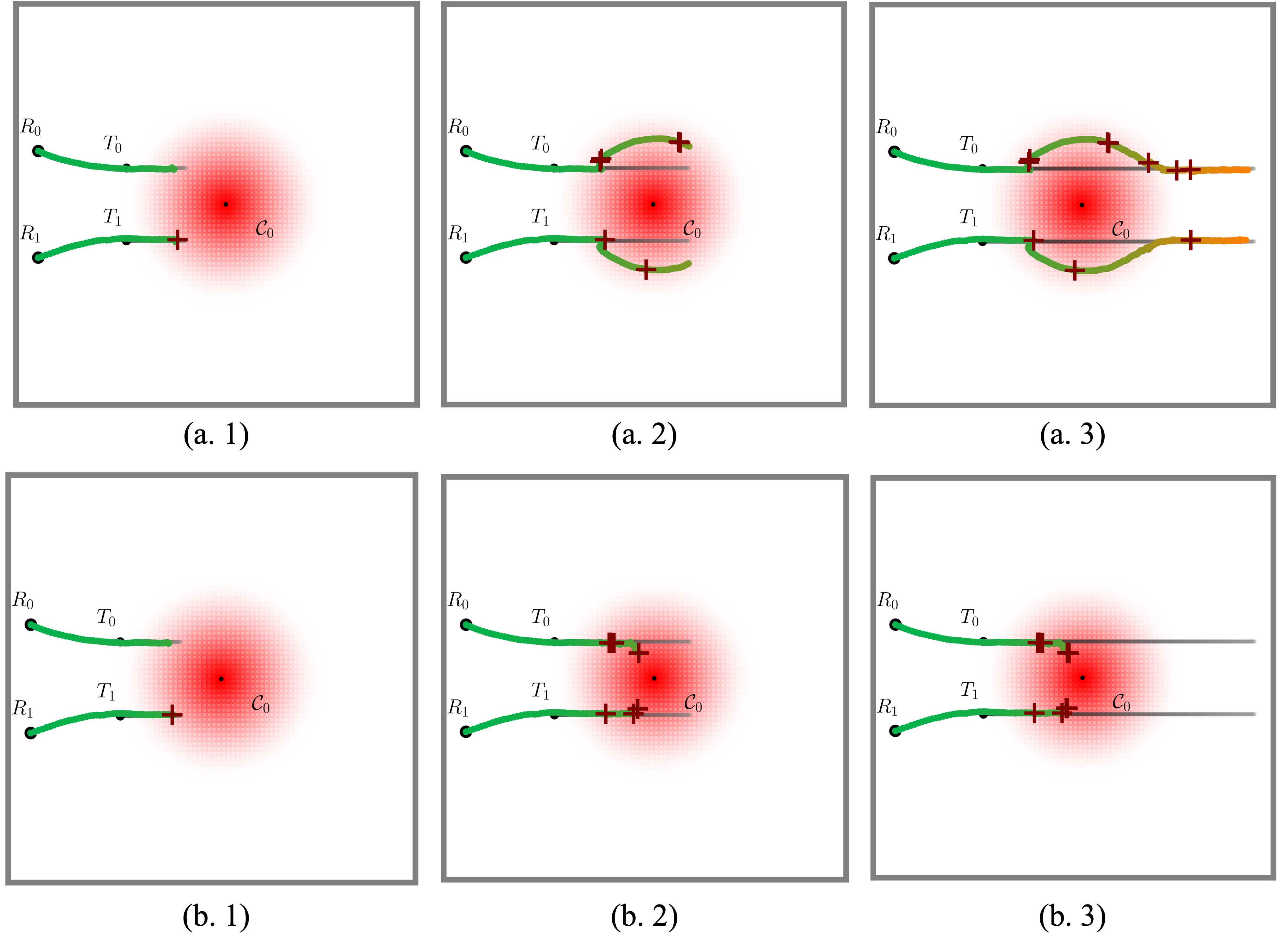}
    \vspace{-0.4cm} 
    \caption{Tracking with a sensing danger zone. The mean position for the center of the danger zone is at [0.1, 0] with a covariance of 0.3. Figures (a.1) - (a.3) are the trajectories with resilient tracking, and (b.1) - (b.3) are those with the vanilla tracking. }
    % \vspace{-0.5cm} 
    \label{fig: sensing_attack}
\end{figure}

\begin{figure}[htp]
\centering
\vspace{-0.00cm}
\subfigure[]{
\includegraphics[height=3.3cm]{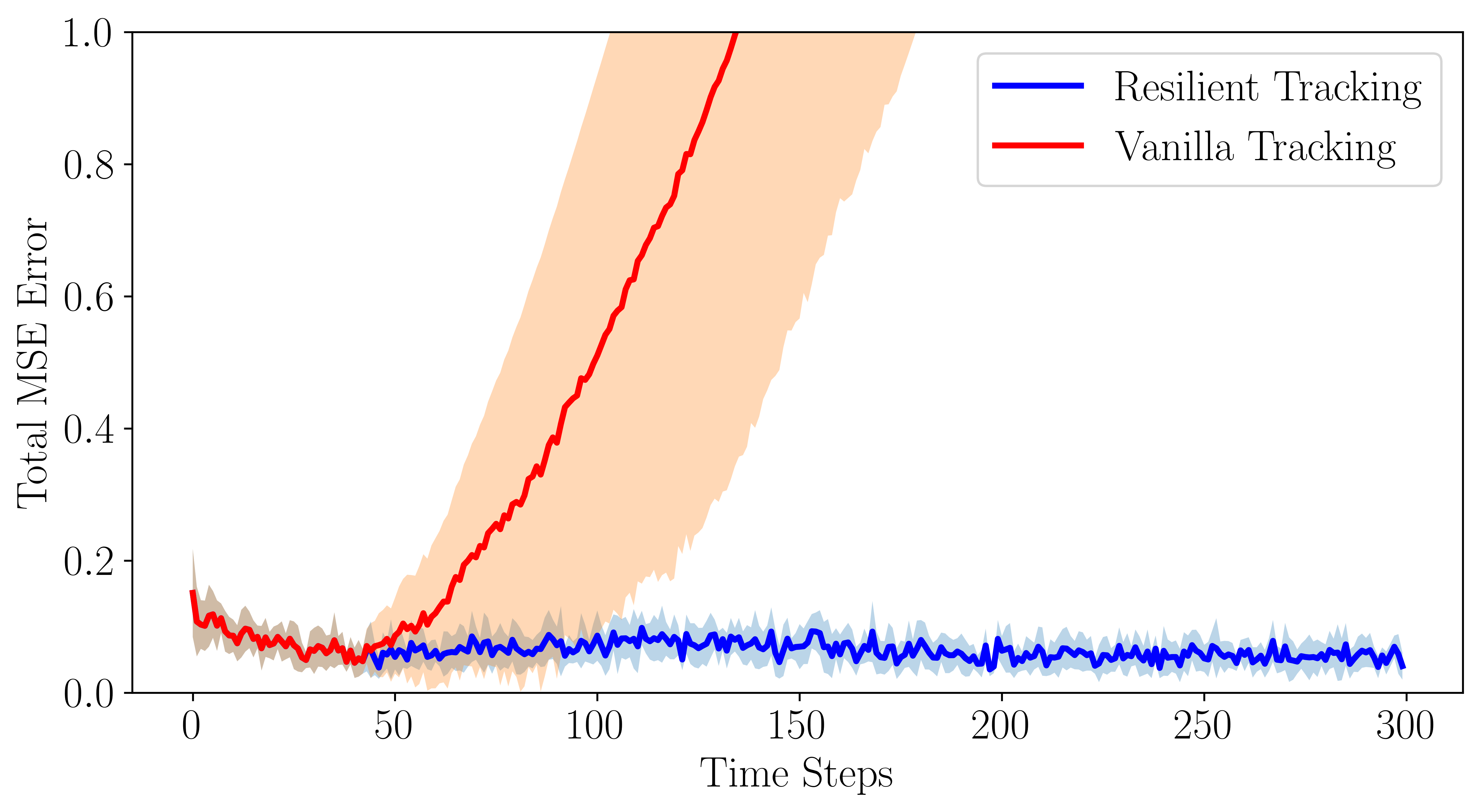}
}
\vspace{-0.38cm}

\subfigure[]{
\includegraphics[height=3.3cm]{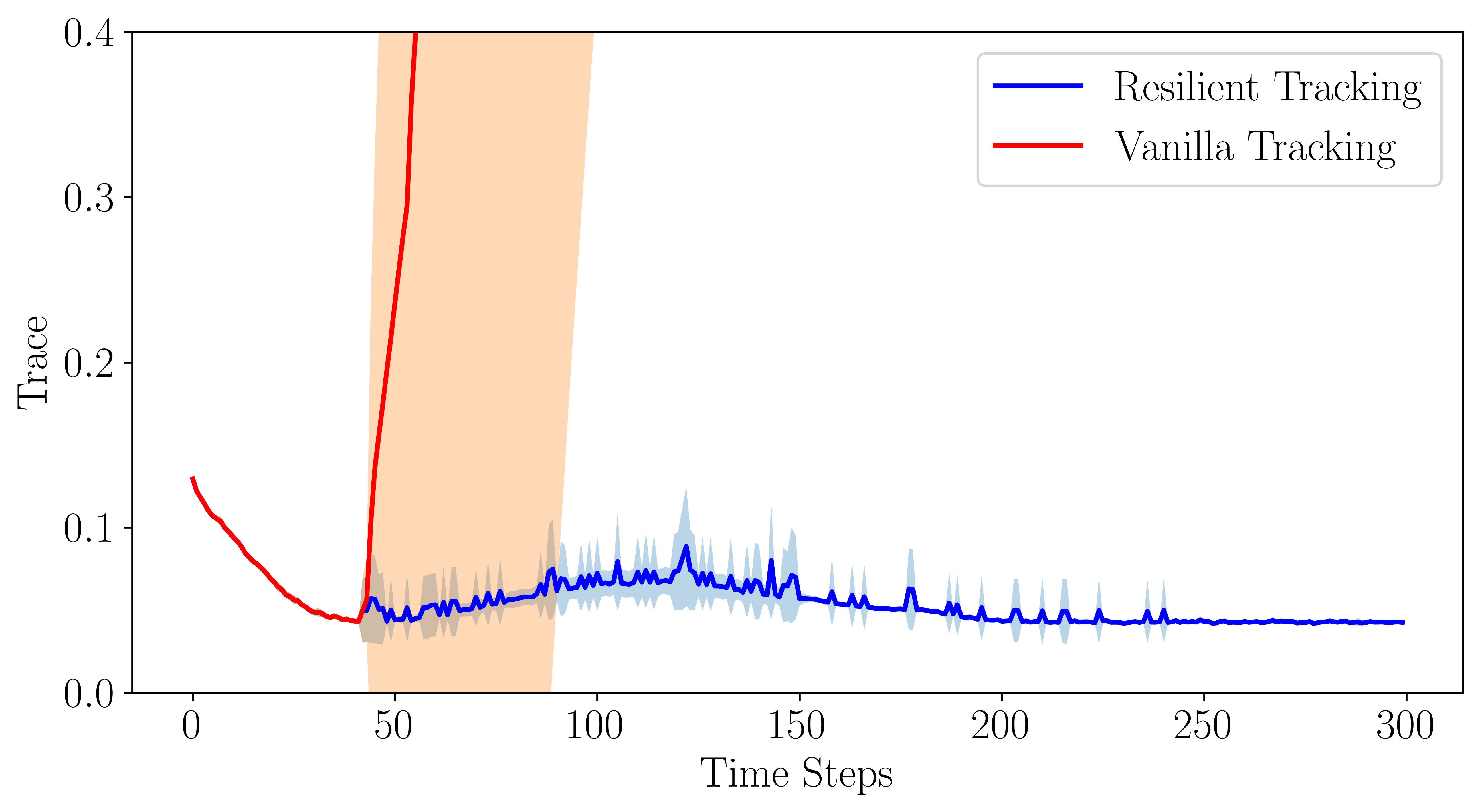}
}
\vspace{-0.2cm}
\caption{Performance of the proposed resilient tracking (blue) versus vanilla tracking (red) under the conditions in Fig.~\ref{fig: sensing_attack}, shown in MSE (a) and the trace of the covariance matrix (b) of the targets' state estimations. The solid line denotes mean values with shaded regions as the $\pm$1 standard deviation.}
\label{fig: sensing_attack_quant}
\vspace{-0.2cm}
\end{figure}

We demonstrate the scenario of two robots, $\mathit{R}_0$ and $\mathit{R}_1$, tracking two targets, $\mathit{T}_0$ and $\mathit{T}_1$, with an unknown sensing danger zone, $\mathit{C}_0$.
We set $\delta_1$ as 0.1 and the attack frequency of the sensing danger zone as 1 Hz.
We perform sampling every 0.1 seconds with a maximum of 300 steps.
The trajectories and attack status from our resilient tracking and vanilla tracking without resiliency to attacks are shown in Fig.~\ref{fig: sensing_attack}. 
Initially, robots have no information about the sensing danger zone and focus on tracking the targets. Then
$\mathit{R}_1$ encounters an attack from the danger zone and immediately broadcasts information about this danger zone to $\mathit{R}_0$.
Notably, the robots can still be attacked even after they have discovered a danger zone. 
The vanilla tracking method suffers from sensing attacks as shown in (b.1) - (b.3) from Fig.~\ref{fig: sensing_attack}; the two robots become immobilized when both of them lose their measurements and cannot continue the target tracking mission. On the other hand, our resilient framework in (a.1) - (a.3) can efficiently identify, broadcast, and recover from the sensing attack.
Fig.~\ref{fig: sensing_attack_quant} demonstrates the mean tracking error and trace of the covariance matrix over ten tests for both the resilient and vanilla tracking methods. Our proposed method maintains high tracking accuracy even in the presence of sensor attacks, whereas the vanilla tracking method fails, resulting in error and trace explosion.

\subsubsection{Tracking under Communication Attack}

\begin{figure}[!ht]
    \centering
    \includegraphics[width=1.0\columnwidth]{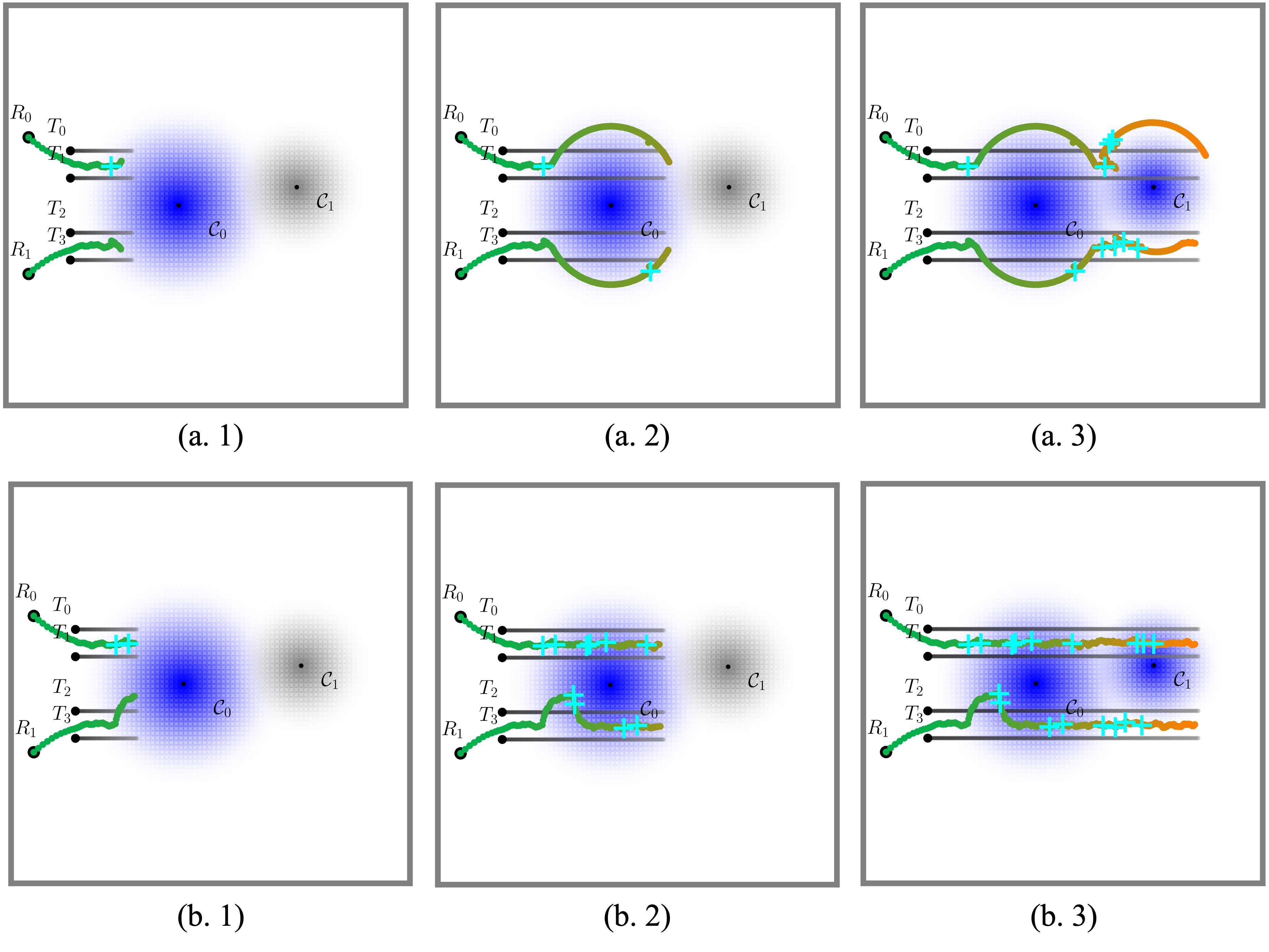}
    \caption{Tracking with two communication danger zones. The first danger zone center point has its mean [-0.3, 0.0] and covariance $\Sigma = 0.3$, the second one centered at  [1.0, 0.2] with $\Sigma = 0.2$. (a.1) - (a.3) show the trajectories from the resilient tracking, and (b.1) - (b.3) show the trajectories from the vanilla tracking. }
    \vspace{-0.3cm} 
    \label{fig: comm_attack}
\end{figure}

\begin{figure}[htp]
\centering
\subfigure[]{
\includegraphics[height=3.3cm]{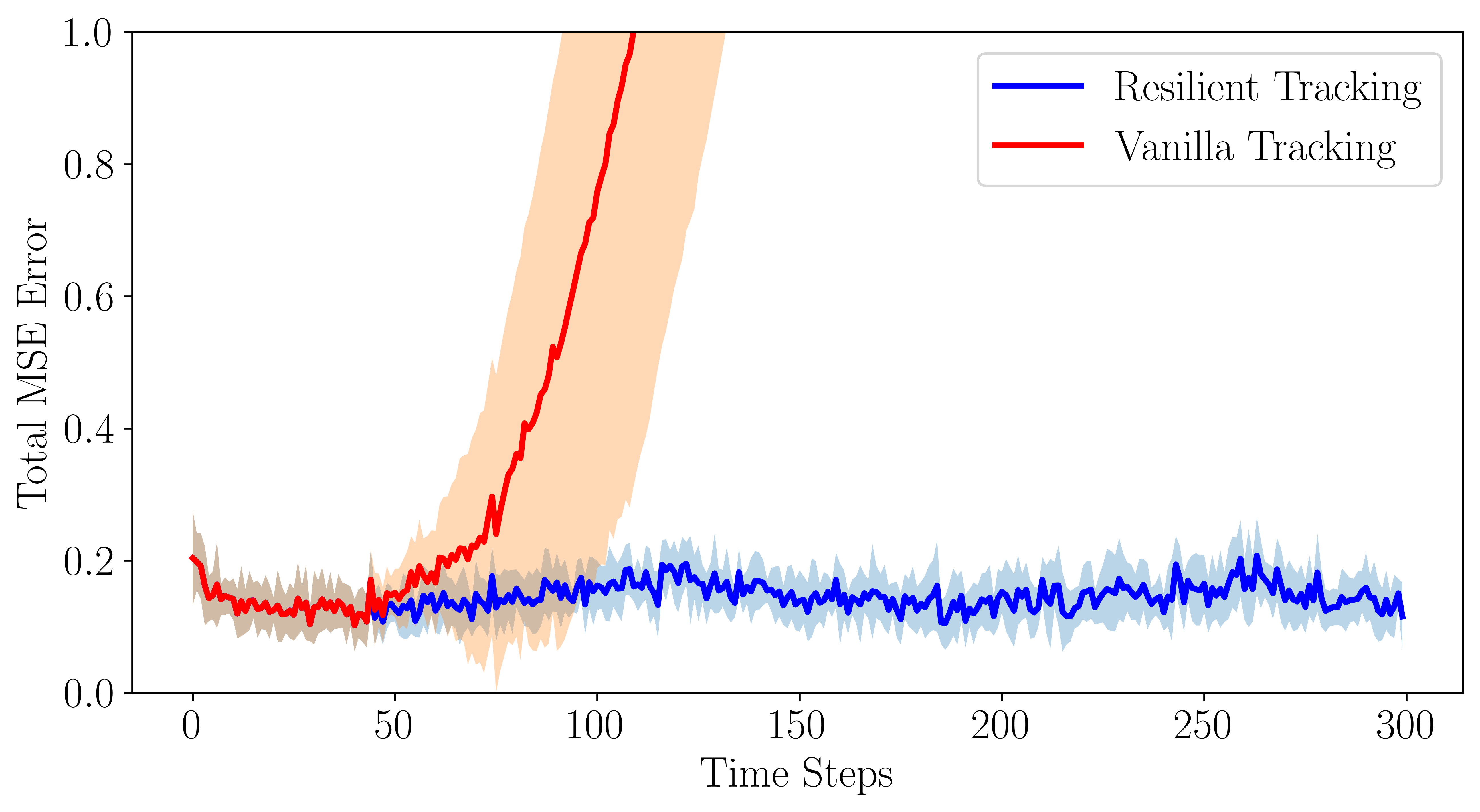}
}
\subfigure[]{
\includegraphics[height=3.3cm]{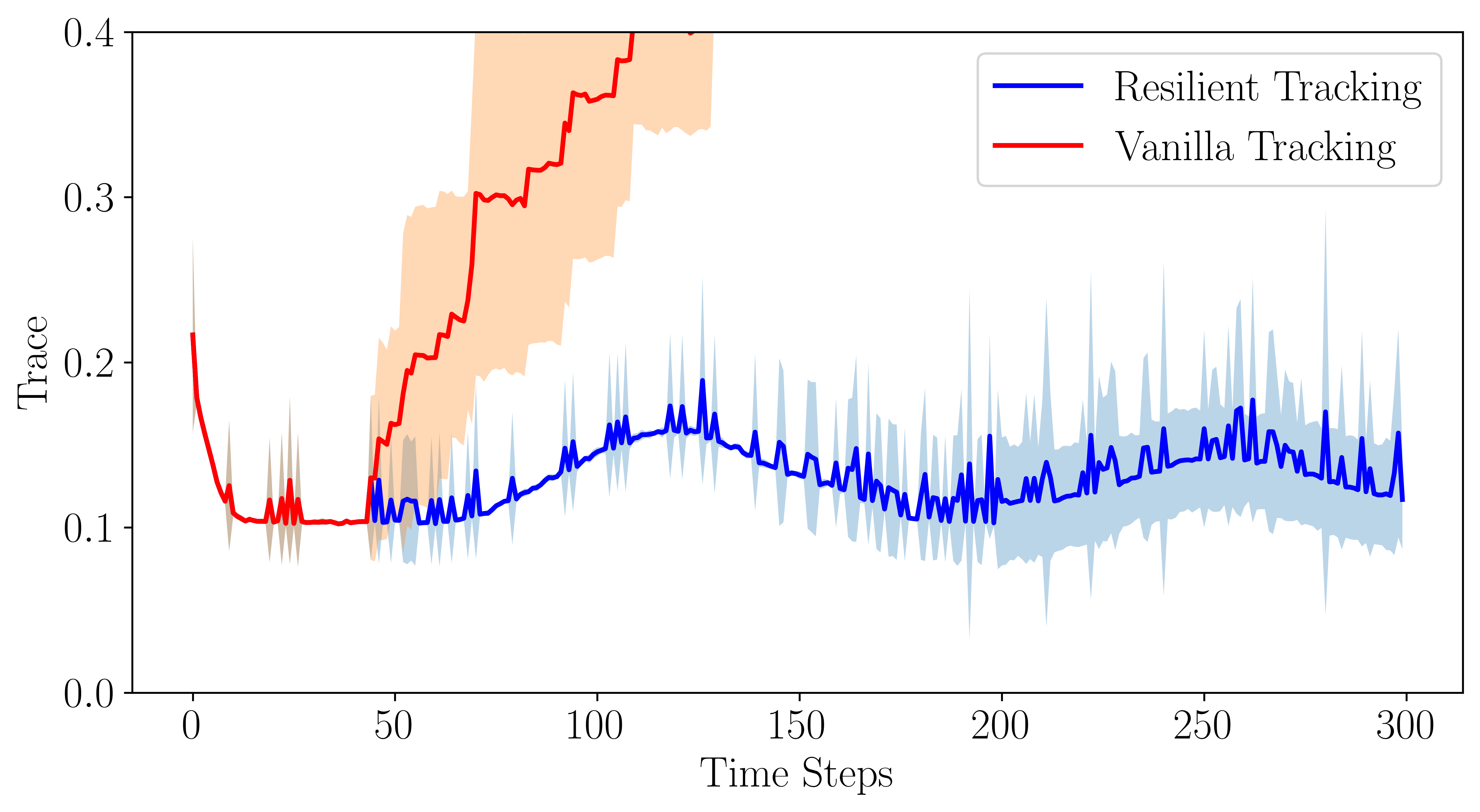}
}
\caption{Performance of the proposed resilient tracking (blue) versus vanilla tracking (red) under the conditions in Fig.~\ref{fig: comm_attack}, shown in MSE (a) and the trace of the covariance matrix (b) of the targets' state estimations.}
\vspace{-0.4cm}
\label{fig: comm_attack_quant}
\end{figure}

We use a similar setup for the communication danger zone to illustrate the performance gap between our resilient tracking and the vanilla tracking with communication attacks.
As shown in Fig.~\ref{fig: comm_attack}, we consider two robots tracking four targets with different parameter setups, and each robot is assigned to track two targets accordingly. 
When robots are close to the communication danger zone, $\mathit{R}_0$ gets attacked and loses its communication link to $\mathit{R}_1$. 
$\mathit{R}_0$ prioritizes escaping from the danger zone and recovering from the communication attack. 
Then, it shares information about the danger zone with $\mathit{R}_1$ so that the latter avoids the danger zone. 
However, for the vanilla tracking, the robots do not share information, and $\mathit{R}_1$ still tries to enter the danger zone after $\mathit{R}_0$ recovers from the communication attack.
Fig.~\ref{fig: comm_attack_quant} shows similar results that the vanilla tracking performs poorly when receiving communication attacks from the environment compared with our resilient tracking.


\subsubsection{Tracking under Multiple Attacks}
\begin{figure*}[!ht]
      \centering
      \vspace{-0.0cm}
      \subfigure[Resilient tracking]{
\includegraphics[width=1.8\columnwidth]{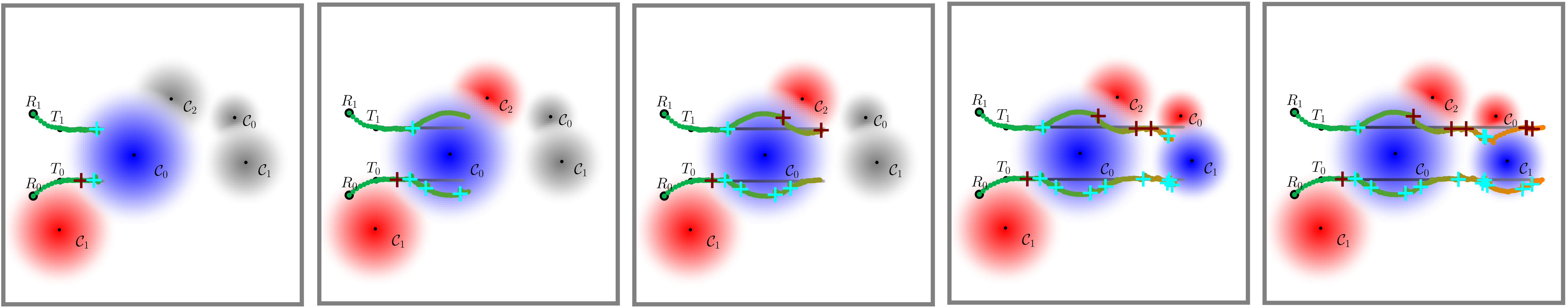}}
\vspace{-0.2cm}
\subfigure[Vanilla tracking]{
\includegraphics[width=1.8\columnwidth]{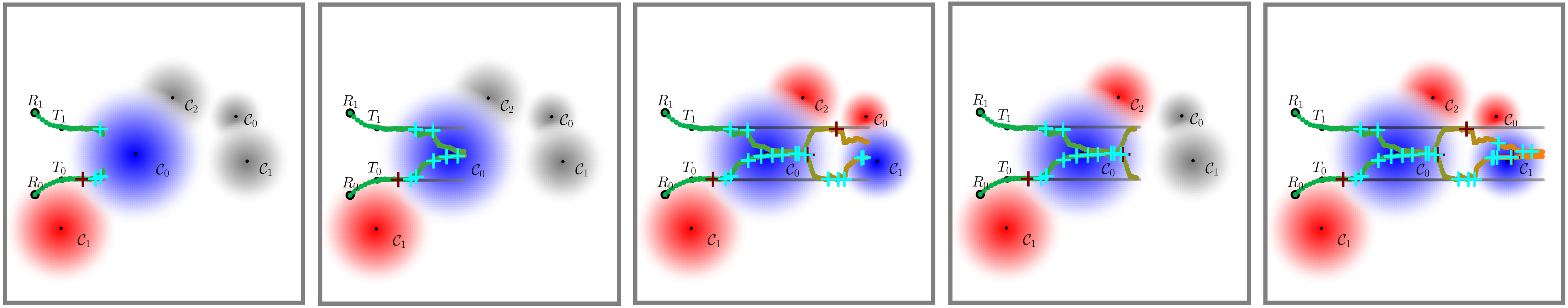}
}
      \caption{Snapshot of two robots tracking two targets with three sensing and two communication danger zones. The sampling time is 0.2 s, with a maximum of 300 steps. Figures from left to right demonstrate the process of discovering the unknown danger zones over time. The robot $\mathit{R}_0$ experiences a sensing attack in sensing danger zone 1, and then successfully recovers. Subsequently, $\mathit{R}_1$ is subjected to communication attacks in a communication danger zone. This process simulates multiple attacks and recoveries across both types of danger zones while gathering critical information about each zone.} 
      \label{fig: combined-attack}
      \vspace{-0.3cm}
\end{figure*}
\begin{figure}[htp]
    \centering
    \subfigure[]{
    \includegraphics[height=3.3cm]{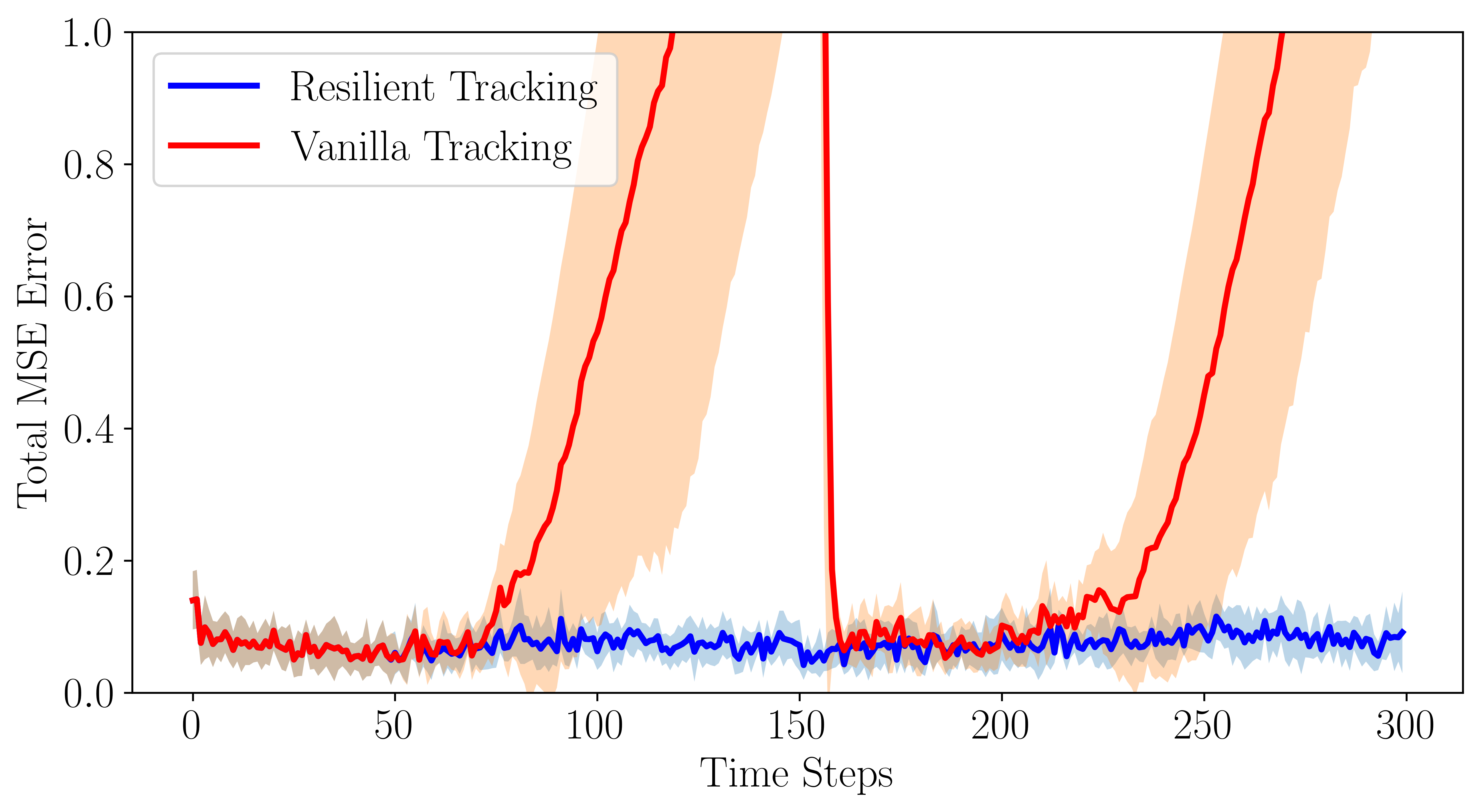}
    }
    \vspace{-0.2cm}
    \subfigure[]{
    \includegraphics[height=3.3cm]{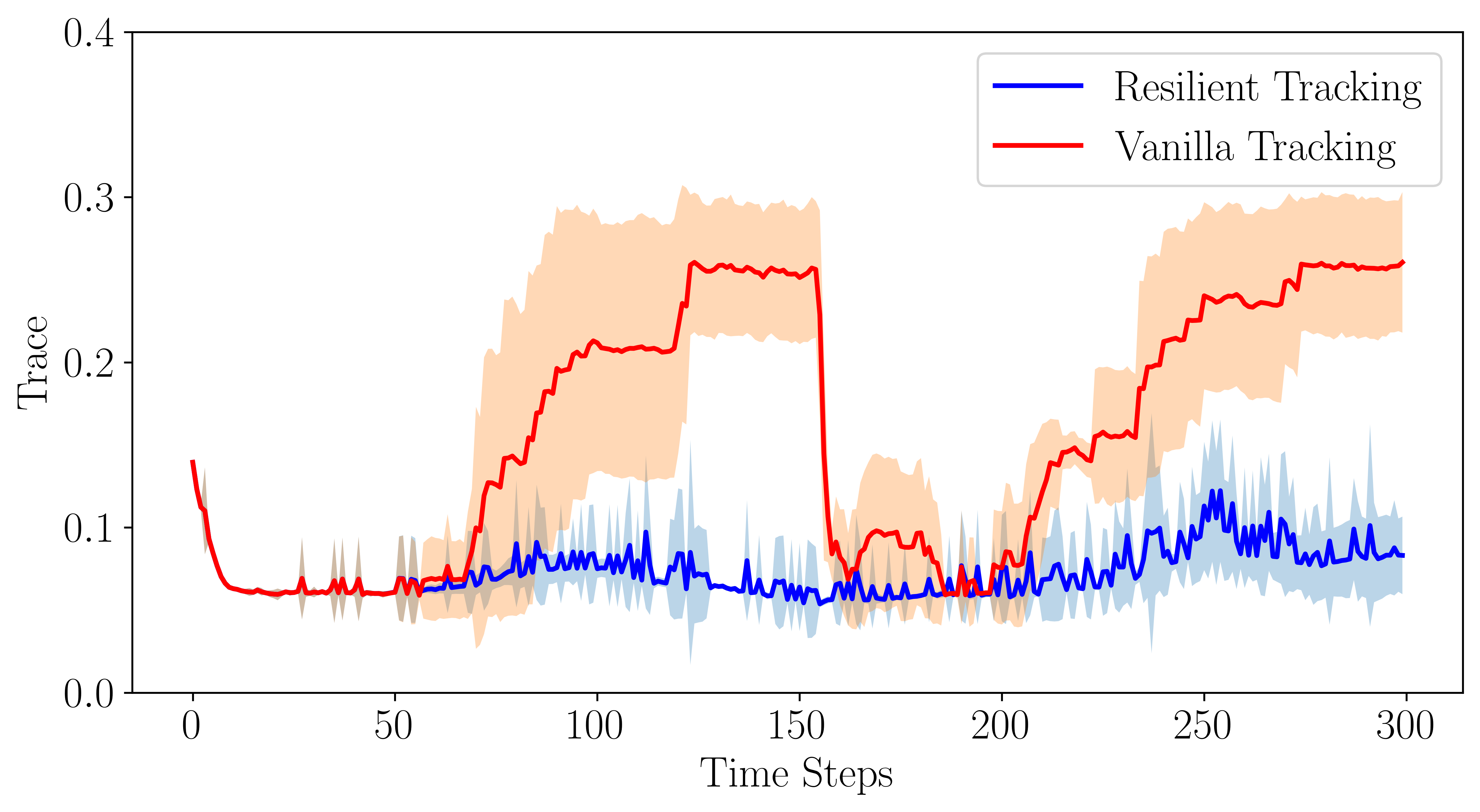}
    }
    \vspace{-0.0cm}
    \caption{Performance of the proposed resilient tracking (blue) versus vanilla tracking (red) under the conditions in Fig.~\ref{fig: combined-attack}, shown in MSE (a) and the trace of the covariance matrix (b) of the targets' state estimations.}
    \vspace{-0.6cm}
    \label{fig: combined-attack-quant}
\end{figure}

We consider a more complicated scenario involving two robots tracking two targets with multiple sensing and communication danger zones in the environment.
As shown in Fig.~\ref{fig: combined-attack}, the two robots can efficiently track the targets while balancing the importance of tracking quality and safety. The danger zones are gradually revealed to the team as the individual robots receive attacks and recover. 
When the robot passes through the overlapping area of two different types of danger zones, a more aggressive strategy enables it to navigate the areas and successfully complete its tracking task.
For vanilla tracking, when both robots lose communication abilities, they converge toward the center to maximize individual target coverage. In contrast, the resilient method enables robots to avoid the danger zone after revealing it. However, this central convergence in the vanilla tracking increases the risk of communication attacks.
Fig.~\ref{fig: combined-attack-quant} shows that the tracking quality from the vanilla tracking suffers from attacks. Although both robots escaped from the attacks towards the middle of the tracking process, and the tracking quality improved, the following attacks deteriorated the tracking quality later in the process. On the other hand, our proposed resilient tracking framework provides consistent tracking performance under a sequence of attacks.

\subsection{Comparison with individual resilient tracking}
With individual resilient tracking, when a robot encounters an attack in a danger zone, it escapes but does not share information with the rest of the team, causing delayed avoidance of danger zones for other robots. In contrast, our proposed resilient tracking enables a robot to share information after recovery, enhancing collaborative tracking performance.

\begin{figure}[!ht]
      \centering
      \subfigure[]{
      \includegraphics[width=0.46\columnwidth]{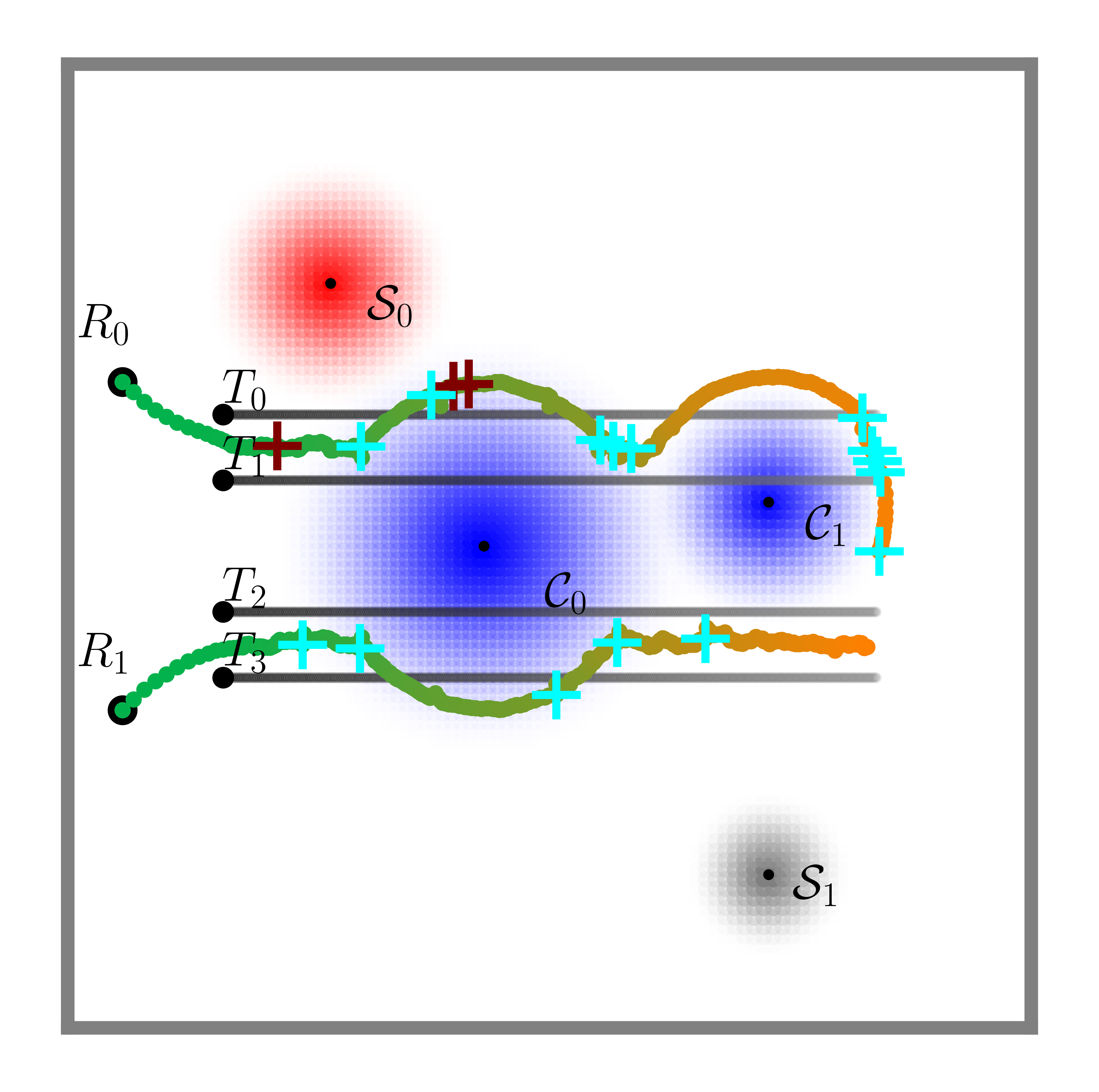}
      }
      \subfigure[]{
      \includegraphics[width=0.46\columnwidth]{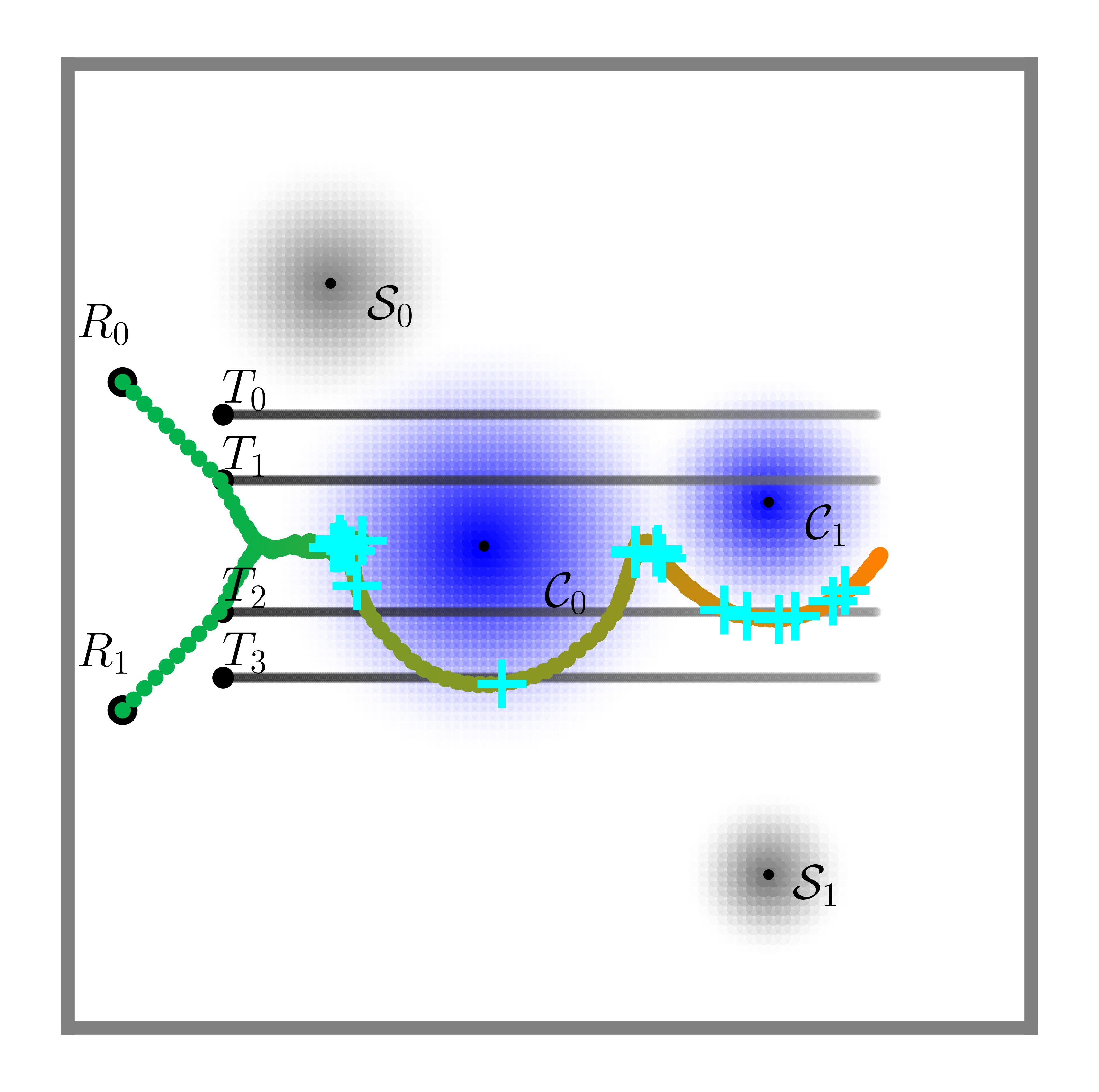}
      }
      \vspace{-0.2cm} 
      \caption{Simulation of two robots tracking two targets under sensing and communication attacks with (a) the proposed collaborative resilient tracking and (b) the individual resilient tracking.} 
      \vspace{-0.3cm} 
      \label{fig: trace-benchmark}
\end{figure}

We benchmark our collaborative resilient tracking against individual resilient tracking to validate the effectiveness of our proposed framework, as shown in Fig.~\ref{fig: trace-benchmark}. In the case of individual resilient tracking, robots cluster in the center due to the lack of information exchange about revealed danger zones, leading to poor tracking performance. In contrast, with our collaborative resilient tracking, robots share knowledge of danger zones after recovery, allowing them to maintain strong tracking performance throughout the process. 

\begin{figure}[!ht]
    \centering
    \subfigure[]{
\includegraphics[height = 3.1cm]{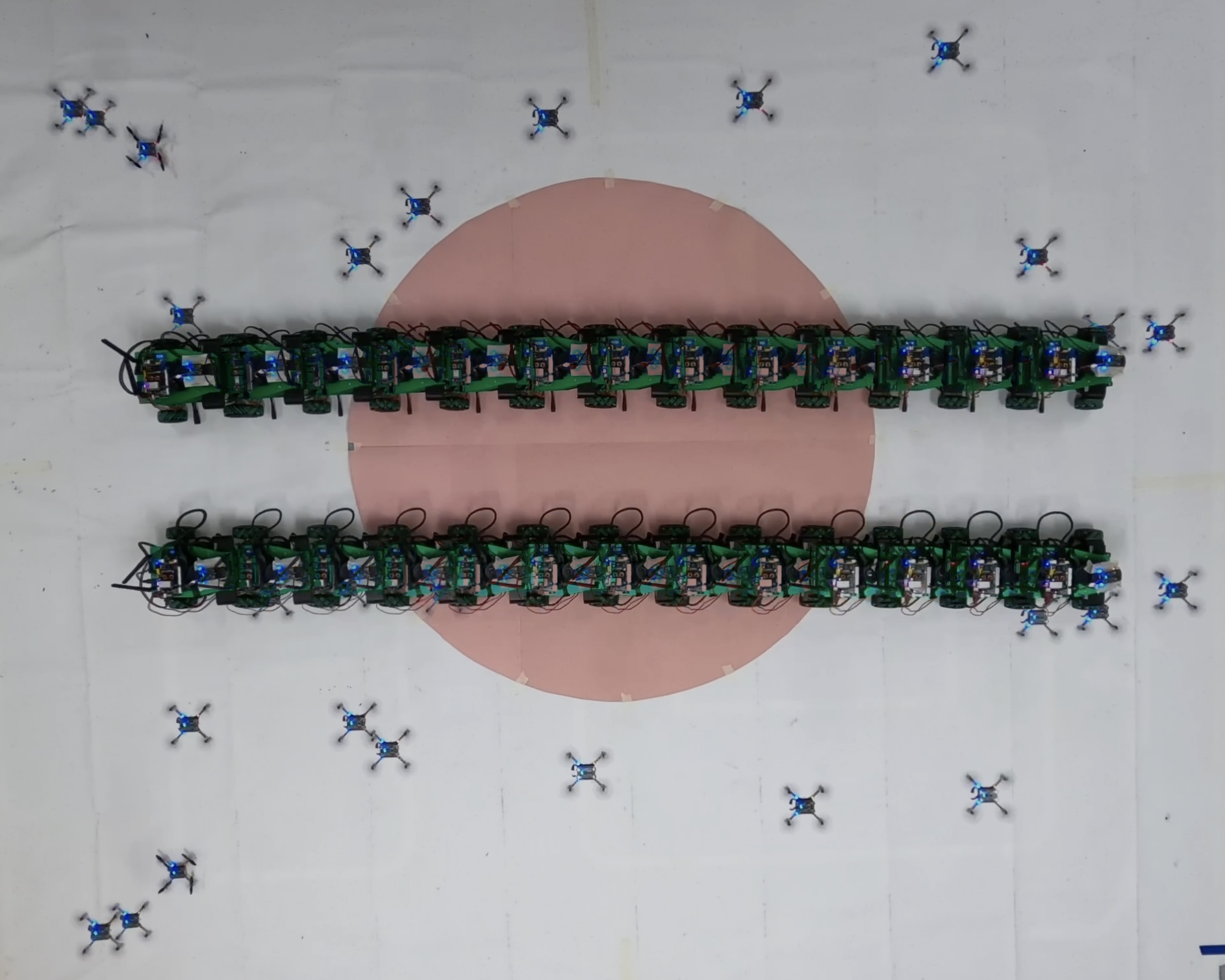}
    }
    \subfigure[]{
\includegraphics[height = 3.1cm]{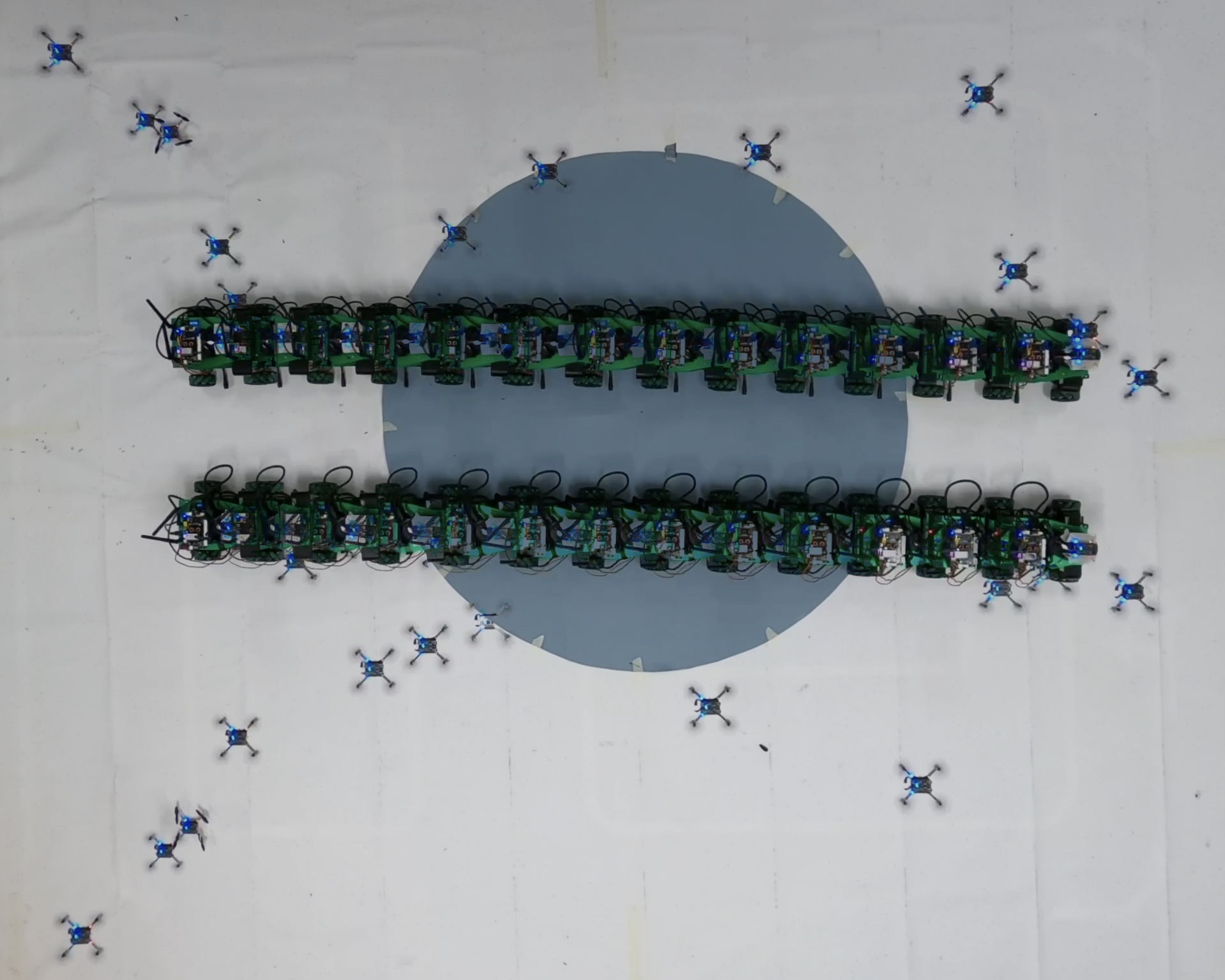}
    }
    \vspace{-0.0cm}  
    \caption{Hardware experiments of two robots tracking two targets with one sensing danger zone (a), and one communication danger zone (b).}
     \label{fig: exp_simple}
    \vspace{-0.0cm}  
\end{figure}

\subsection{Hardware Experiments}

We validate the resiliency of our proposed framework through multiple hardware experiments conducted in various initially unknown hazardous environments. Following a similar experimental setup as in our prior work~\cite{liu2024multi}, we use Crazyflie drones~\cite{crazyflie} as trackers and Yahboom ROSMASTER X3 ground robots~\cite{yahboom} as targets. The scenario involving a single danger zone is illustrated in Fig.~\ref{fig: exp_simple}, while the case with multiple danger zones is shown in Fig.~\ref{fig: fig1}. In both scenarios, our resilient tracking framework enables robots to resiliently react and recover when one of them is attacked and identifies a danger zone. In the case of a sensing danger zone, all robots receive the danger zone information simultaneously and immediately escape, as they can share information in real time. However, in a communication danger zone, the response is sequential, where one robot escapes and recovers first before sharing the information, allowing the next robot to avoid the zone accordingly. This strategy ensures effective tracking performance even in the presence of attacks and environmental uncertainties.

\section{Conclusion}
 
We present a resilient multi-robot multi-target tracking architecture under unknown sensing and communication danger zones. 
We address this multi-robot planning problem through an optimization formulation that allows robots to adjust the prioritization of task goals, considering factors such as tracking performance, safety, and efficiency.
Simulations and real robot experiments show the resiliency of our proposed framework when encountering sensing and/or communication attacks. In the future, we will extend the resilient framework to more complex and realistic scenarios, including real-time perception, restricted resources, limited sensing and communication ranges, dynamic danger zones, and adversarial targets.

\bibliography{references}

\end{document}